\documentclass{ecai}

\usepackage{microtype}
\usepackage{graphicx}
\usepackage{subfigure}
\usepackage{booktabs}
\usepackage{hyperref}
\usepackage{amsmath}
\usepackage{amssymb}
\usepackage{mathtools}
\usepackage{amsthm}
\usepackage{amsfonts}       % blackboard math symbols
\usepackage{amssymb}
\usepackage{microtype}      % microtypography
\usepackage{nicefrac}       % compact symbols for 1/2, etc.
\usepackage{pifont}
\usepackage{xcolor}         % colors
\usepackage{tablefootnote}

\usepackage{multirow}
\usepackage{multicol} % to delete
\usepackage{latexsym}

\ecaisubmission   % inserts page numbers. Use only for submission of paper.
\def\our{TreeFlow}

\newcommand{\x}{\mathbf{x}}

\newcommand{\y}{\mathbf{y}}
\newcommand{\D}{\mathcal{D}}
\newcommand{\Rs}{\mathcal{R}}
\newcommand{\R}{\mathbb{R}}
\newcommand{\1}{\mathbf{1}}

\newcommand{\cmark}{\ding{51}}%
\newcommand{\xmark}{\ding{55}}%

\begin{document}

\begin{frontmatter}

\title{TreeFlow: Going Beyond Tree-based Parametric Probabilistic Regression}

\author[A, B]{\fnms{Patryk}~\snm{Wielopolski}\thanks{Corresponding Author. Email: patryk.wielopolski@pwr.edu.pl}}
\author[A, C]{\fnms{Maciej}~\snm{Zięba}}

\address[A]{Wrocław University of Science and Technology}
\address[B]{DataWalk}
\address[C]{Tooploox}

\begin{abstract}
The tree-based ensembles are known for their outstanding performance in classification and regression problems characterized by feature vectors represented by mixed-type variables from various ranges and domains. However, considering regression problems, they are primarily designed to provide deterministic responses or model the uncertainty of the output with Gaussian or parametric distribution. In this work, we introduce \our{}, the tree-based approach that combines the benefits of using tree ensembles with the capabilities of modeling flexible probability distributions using normalizing flows. The main idea of the solution is to use a tree-based model as a~feature extractor and combine it with a conditional variant of normalizing flow. Consequently, our approach is capable of modeling complex distributions for the regression outputs. We evaluate the proposed method on challenging regression benchmarks with varying volume, feature characteristics, and target dimensionality. We obtain the SOTA results for both probabilistic and deterministic metrics on datasets with multi-modal target distributions and competitive results on unimodal ones compared to tree-based regression baselines.
\end{abstract}

\end{frontmatter}

\section{Introduction}

The modern tree-based models achieve outstanding results for problems where the data representation is tabular, the number of training examples is limited, and the input feature vector is represented by mixed-type variables from various ranges and domains. Most of such algorithms focus on providing deterministic predictions, paying no attention to the probabilistic nature of the provided output. However, for many practical applications, it is impossible to deliver an exact target value based on the given input factors. Consider the regression problem of predicting the future location of the vehicle that is approaching a roundabout \cite{zikeba2020regflow}. Having past coordinates and other information aggregated in current and past states, we cannot unambiguously predict which of the three remaining exits from the roundabout will be taken by the tracked object. Therefore, it is more beneficial to provide multimodal probability distribution for future locations instead of a single deterministic prediction oscillating around one mode.

Due to the tractable closed-form, the standard approaches assume to model regression uncertainty using Gaussian or parametric distributions \cite{murphy2012machine}. The well-known deterministic gradient boosting machine method adopted those approaches to tree-based structures \cite{DuanADTBNS20, MalininPU21, sprangers2021probabilistic}. Consequently, they can capture the uncertainty of the regression outputs with a standard family of distributions. 

The major limitation of the current approaches is modeling regression outputs using only Gaussians. Moreover, it is not trivial to extend them to a mixture of Gaussians to capture the multi-modalities of the predictions. Creating multivariate extensions of such models is also ineffective, especially for higher dimensionality, due to the need to estimate the complete covariance matrix. 

To reduce the limitations of existing methods, we introduce \our{} - a novel tree-based approach for modeling probabilistic regression. The proposed method combines the benefits of using tree-based structures as feature extractors with the normalizing flows \cite{rezende2015variational} capable of modeling flexible data distributions. We introduce the novel concept of combining forest structure with a conditional flow variant to model uncertainty for regression output. Thanks to that approach, we can model complex non-Gaussian or in general non-parametric data distributions even for high-dimensional predictions. We confirm the quality of the proposed model in the experimental part, where we show the superiority of our method over the baselines.

To summarize, our contributions are as follows:

\begin{itemize}
    \item According to our knowledge, for the first time, tree-based models are used to model non-parametric probabilistic regression for both uni- and multi-variate predictions.
    \item We propose a novel approach for combining tree-based models with conditional flows via binary representation of the forest structure.
    \item We obtain the SOTA results for both probabilistic (NLL, CRPS) and deterministic (RMSE) metrics on datasets with multi-modal target distributions and competitive results on unimodal ones compared to tree-based regression baselines.
\end{itemize}

\section{Background}

Assume we have a dataset $\D=\{(\x_n,\y_n)\}_{n=1..N}$ where $\x_n = (x_n^1, \ldots, x_n^D)$ is a $D$-dimensional random vector of features and $\y_n = (y_n^1, \ldots, y_n^P)$ is a $P$-dimensional vector of targets. We consider regression problems, thus, we assume that $y_n^p\in \R$. Additionally, when $P=1$ we will refer to that as a univariate regression problem, and when $P\geq2$ as a multivariate regression problem.

For the probabilistic regression task, we aim at modeling conditional probability distribution $p(\mathbf{y}|\mathbf{x})$. Assuming some parametrization of the regression model $\boldsymbol{\theta}$, the problem of training the probabilistic model can be expressed as minimisation of the conditional negative log likelihood function (NLL) given by $Q(\boldsymbol{\theta})=-\sum_{n=1}^N\log{p(\y_n|\x_n, \boldsymbol{\theta})}$. During the training procedure we aim at finding $\boldsymbol{\theta}^*=\arg \min_{\boldsymbol{\theta}}Q(\boldsymbol{\theta})$.

\paragraph{Decision Tree Ensembles}
Decision Tree \cite{Breiman1983ClassificationAR} recursively partition feature space $\R^D$ into $K$ disjoint regions $\Rs_k$ (tree leaves) and for each region assign value $w_k$. Formally, the model can be written as $ h(\x) = \sum_{k=1}^{K} w_k \1_{\{\x \in \Rs_k \}}$.

Decision Tree Ensembles are constructed of multiple, usually shallow decision trees, whose results are differently aggregated depending on the training mode. In general, we distinguish two main approaches: independent model training with average or majority voting such as Random Forest \cite{breiman2001random}, and iterative model training with additive aggregation such as Gradient Boosting Machine (GBM) \cite{friedman2001greedy}. 

For the univariate probabilistic regression, GBM optimizes the loss function given by negative log likelihood (NLL). Then it assumes the target variable has a Gaussian distribution, i.e.,
\begin{equation}
    p(y |\x , \boldsymbol{\theta}^{(t)}) = \mathcal{N}(y|\mu^{(t)},\sigma^{(t)}),
    \label{eq:prob_cat}
\end{equation}
where $\{\mu^{(t)},\log\sigma^{(t)}\} = F^{(t)}(\x)$ and $F^{(t)}(\x)$ is an output of $t$-th tree from GBM model consisted of $T$ trees. In the multivariate case, it assumes Multivariate Normal distribution and uses the parametrization trick with Cholesky decomposition of the covariance matrix which reduces the number of parameters. 

In practice, NGBoost \cite{DuanADTBNS20} supports both uni- and multi-variate Gaussian distributions and estimates each parameter using one underlying model. CatBoost \cite{MalininPU21} supports only univariate Gaussians but estimates all distribution parameters using only one model. Moreover, it provides deterministic multivariate regression with the same property, which keeps the total number of trees relatively small. In this case, the loss function is Multioutput Root Mean Squared Error (MultiRMSE).

%\begin{equation}
%    \sqrt{\frac{1}{N}\displaystyle\sum\limits_{n=1}^{N}\sum\limits_{p=1}^{P}(\hat{y}_{n}^{p} - y_{n}^{p})^{2}}\,,
%\label{eq:multirmse}
%\end{equation}
%where $\hat{y}_{n}^{p}$ is an output of the model for the $n$th example for the $p$th dimension.

\paragraph{Normalizing Flows}

Normalizing flows \cite{rezende2015variational} represent the group of generative models that can be efficiently trained via direct likelihood estimation thanks to the application of a change-of-variable formula. Practically, they utilize a series of (parametric) invertible functions: $\mathbf{y}=\mathbf{f}_n \circ \dots \circ \mathbf{f}_1(\mathbf{z})$. Assuming given base distribution $p(\mathbf{z})$ for $\mathbf{z}$, the log likelihood for $\mathbf{y}$ is given by $\log p(\mathbf{y}) = \log p(\mathbf{z}) - \sum_{n=1}^N \log  \left| \det \frac{\partial \mathbf{f}_n}{\partial \mathbf{z}_{n-1}} \right|$. In practical applications $p(\mathbf{y})$ represents the distribution of observable data and $p(\mathbf{z})$ is usually assumed to be Gaussian with independent components. 

The sequence of discrete transformations can be replaced by continuous alternative by application of Continuous Normalizing Flows (CNFs) \cite{ChenRBD18, grathwohl2018ffjord} where the aim is to solve the differential equation of the form $\frac{\partial \mathbf{z}}{\partial t}=\mathbf{g}_{\boldsymbol{\beta}}(\mathbf{z}(t), t)$, where $\mathbf{g}_{\boldsymbol{\beta}}(\mathbf{z}(t), t)$ represents the function of dynamics, described by parameters $\boldsymbol{\beta}$. Our goal is to find solution of the equation in $t_1$, $\mathbf{y}:=\mathbf{z}(t_1)$, assuming the given initial state $\mathbf{z}:=\mathbf{z}(t_0)$ with a known prior. The transformation function $\mathbf{f}_{\boldsymbol{\beta}}$ is defined as:

\begin{equation}
    \mathbf{y} = \mathbf{f}_{\boldsymbol{\beta}}( \mathbf{z} ) =  \mathbf{z} + \int^{t_1}_{t_0} \mathbf{g}_{\boldsymbol{\beta}}(\mathbf{z}(t), t) dt.
\label{eq:f}
\end{equation}

The inverted form of the transformation can be easily computed using the formula: $    \mathbf{f}_{\boldsymbol{\beta}}^{-1}( \mathbf{y})  = \mathbf{y} - \int^{t_1}_{t_0} \mathbf{g}_{\boldsymbol{\beta}}(\mathbf{z}(t), t) dt$. The log-probability of $\mathbf{y}$ can be computed by:
\begin{equation}
    \log p(\mathbf{y}) = \log p( \mathbf{f}_{\boldsymbol{\beta}}^{-1}( \mathbf{y}) ) - \int^{t_1}_{t_0} \mathrm{Tr} \left( \frac{\partial \mathbf{g}_{\boldsymbol{\beta}}}{\partial \mathbf{z}(t)} \right) dt,
\end{equation}
where $\mathbf{f}_{\boldsymbol{\beta}}^{-1}( \mathbf{y}) = \mathbf{z}.$

CNFs are rather designed to model complex probability distributions for low-dimensional data, what was confirmed in various applications including point cloud generation \cite{yang2019pointflow}, future prediction \cite{zikeba2020regflow} or probabilistic few-shot regression \cite{sendera2021non}. Compared to models like RealNVP \cite{dinh2016density} or Glow \cite{kingma2018glow}, they can be successfully applied to one-dimensional data and achieve better results for tabular datasets.     

\section{\our{}} \label{sec:treeflow}

\begin{figure*}[t]
    
    \begin{center}
    \centerline{\includegraphics[width=0.8\textwidth]{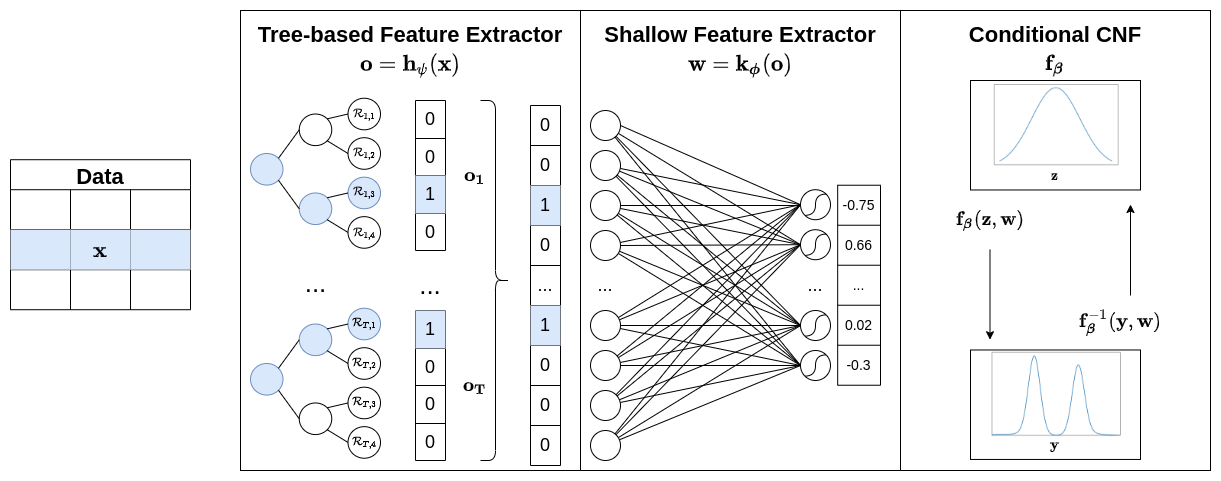}}
    \caption{\our{} architecture. The proposed model consists of three components: Tree-based Feature Extractor, Shallow Feature Extractor, and Conditional CNF module. The role of the first component is to extract the vector of binary features from the structure of the tree-based ensemble model. The Shallow Feature Extractor is a neural network responsible for mapping high-dimensional binary vectors returned by the Tree-based Feature Extractor to low-dimensional feature space. The resulting vector is further passed to the conditional CNF module as a conditioning factor. The goal of the last component is to model complex probability distribution.}
    \label{fig:architecture}
    \end{center}
    
\end{figure*}

Tree-based methods obtain superior results on tabular datasets and have developed multiple techniques to deal with categorical variables, null values, etc. but are limited to distributions with explicitly provided probability distribution functions, e.g., Gaussian. We want to overcome this limitation by introducing \our{} - method for uni- and multi-variate tree-based probabilistic regression with non-Gaussian and multi-modal target distributions. The main idea of the solution is to combine the benefits of using tree ensembles with the capabilities of modeling flexible probability distributions using conditional normalizing flows.

The architecture of \our{} is provided in fig. \ref{fig:architecture}. The proposed model consists of three components: Tree-based Feature Extractor, Shallow Feature Extractor, and conditional CNF module. The role of the first component is to extract the vector of binary features from the structure of the tree-based ensemble model for a given input observation $\mathbf{x}$. The problem of extracting a~unified vector from the tree-based ensemble model is non-trivial due to the complex structure and a large number of base learners.  
Motivated by the fact that crucial information extracted from input examples is stored in the leaves, we propose a binary occurrence representation that is the most lightweight approach assuming thousands of trees.
Formally it could be written as $\mathbf{h}_{\boldsymbol{\psi}}(\x) = [\mathbf{o_1}, \ldots, \mathbf{o_T}]$, where $\mathbf{h}_{\boldsymbol{\psi}}(\x)$ is Tree-based Feature Extractor with parameters $\boldsymbol{\psi}$ and $\mathbf{o_i} = [\1_{\{\x \in \Rs_{i, 1} \}}, \ldots, \1_{\{\x \in \Rs_{i, K} \}} ]$ where $\Rs_{i, k}$ is a region of $k$th leaf of $i$th decision tree in the forest structure.   

The size of vector $\mathbf{o}$ is significantly larger than the size of the regression variable $\mathbf{y}$. If we deliver directly the large sparse binary vector as a CNF conditioning component: (i) the number of CNF parameters grows significantly, (ii) the conditioning component dominates training, and (iii) the ordinary differential equation (ODE) solver slows down significantly and behaves in an unstable way. Therefore, we use an additional Shallow Feature Extractor $\mathbf{k}_{\mathbf{\phi}}(\cdot)$, that is represented by a neural network responsible for mapping high-dimensional binary vector $\mathbf{o}$ returned by $\mathbf{h}_{\boldsymbol{\psi}}(\x)$ to low-dimensional feature representation $\mathbf{w}=\mathbf{k}_{\mathbf{\phi}}(\mathbf{o})$. The low-dimensional representation $\mathbf{w}$ of the sparse embedding $\mathbf{o}$ is further passed to the conditional CNF module as a conditioning factor. We postulate to use the variant of the conditional flow-based model provided in \cite{yang2019pointflow, sendera2021non}, where $\mathbf{w}$ is delivered to the function of dynamics, $\mathbf{g}_{\boldsymbol{\beta}}(\mathbf{z}(t), t, \mathbf{w})$. The transformation function is given by eq. \eqref{eq:f} is represented as:

\begin{equation}
    \mathbf{y} = \mathbf{f}_{\boldsymbol{\beta}}( \mathbf{z}, \mathbf{w}) =  \mathbf{z} + \int^{t_1}_{t_0} \mathbf{g}_{\boldsymbol{\beta}}(\mathbf{z}(t), t, \mathbf{w}) dt.
\label{eq:f_c}
\end{equation}

The inverse form of the transformation $\mathbf{f}_{\boldsymbol{\beta}}(\cdot)$ given the same $\mathbf{w}$ in both directions is simply:
$\mathbf{z} = \mathbf{f}^{-1}_{\boldsymbol{\beta}}( \mathbf{y}, \mathbf{w}) =  \mathbf{y} - \int^{t_1}_{t_0} \mathbf{g}_{\boldsymbol{\beta}}(\mathbf{z}(t), t, \mathbf{w}) dt$. For a given model, we can easily calculate the log-probability of regression output $\mathbf{y}$, given the input $\mathbf{x}$ \cite{grathwohl2018ffjord}:

\begin{scriptsize}
\begin{align}
    \log p(\mathbf{y}|\mathbf{w}) &= \log p( \mathbf{f}_{\boldsymbol{\beta}}^{-1}( \mathbf{y}, \mathbf{w})) - \int^{t_1}_{t_0} \mathrm{Tr} \left( \frac{\partial \mathbf{g}_{\boldsymbol{\beta}}(\mathbf{z}(t), t, \mathbf{w})}{\partial \mathbf{z}(t)} \right) dt,
    \label{eq:logprob}
\end{align}
\end{scriptsize}
where $\mathbf{w}=\mathbf{k}_{\mathbf{\phi}}(\mathbf{o})$, and $\mathbf{o} = \mathbf{h}_{\boldsymbol{\psi}}(\x)$. With the model defined in the following way, we can easily calculate the exact value of log-probability for any possible regression outputs. We can also utilize the generative capabilities of the model by generating samples from a known prior $p(\mathbf{z})$ and transforming them into the space of regression outputs using the function given by eq.~\eqref{eq:f_c}. 

We aim at training the model by optimizing the NLL for a log probability defined by eq. \eqref{eq:logprob} and the set of trainable parameters $\boldsymbol{\theta}=\{\boldsymbol{\psi}, \boldsymbol{\phi}, \boldsymbol{\beta}\}$. In the perfect scenario, we should optimize the entire model in an end-to-end fashion, jointly updating the parameters of the Tree-based Feature Extractor $\boldsymbol{\psi}$, Shallow Feature Extractor $\boldsymbol{\phi}$, and conditional CNF $\boldsymbol{\beta}$. However, the Shallow Feature Extractor needs to have a constant size input which cannot be easily obtained from our Tree-based Feature Extractor as it learns iteratively. To overcome this limitation, we perform two-staged learning.

In the first stage, only the parameters of Tree-based Feature Extractor $\boldsymbol{\psi^*}$ are trained by optimizing the surrogate criterion specific to the type of tree-based architecture. In our work, we utilize the CatBoost model as it out-of-the-box supports categorical features and null values. Therefore, following \cite{MalininPU21} we train the Tree-based feature extractor by optimizing NLL loss for a standard Gaussian regression output given by eq. \eqref{eq:prob_cat}. For the multivariate case, we use the protocol from \cite{ProkhorenkovaGV18} and train the feature extractor by optimizing MultiRMSE.

Given the Tree-based Feature Extractor parameters, we train the remaining components of our model in an end-to-end fashion. Formally, given the estimated parameters $\boldsymbol{\psi^*}$ for $\mathbf{h}_{\boldsymbol{\psi}}(\x)$ we train the model by optimizing NLL with log-probability given by eq.~\eqref{eq:logprob} with respect to remaining parameters $\boldsymbol{\phi}$ and $\boldsymbol{\beta}$ using the standard gradient-based approach. 

The two-stage training has a couple of advantages compared to the end-to-end approach. First, any trained tree-based ensemble can be used as a feature extractor. Second, extracting the forest structure together with optimizing the parameters of the remaining components of the system in an end-to-end fashion is non-trivial and requires handcrafting the training procedure for a particular type of tree-based learner.

\section{Related works}

One of the best-known examples of gradient boosting methods is XGBoost \cite{ChenG16} which iteratively combines weak regression trees to obtain accurate predictions. Further extensions to this method consist of LightGBM \cite{KeMFWCMYL17} and CatBoost \cite{ProkhorenkovaGV18} which introduce multiple novel techniques to obtain even better point estimates. Recently they have been extended to a probabilistic framework to model the whole probability distributions.

One such approach is NGBoost (Natural Gradient Boosting) \cite{DuanADTBNS20} algorithm, which can model any probabilistic distribution with a defined probability density function, e.g., Univariate Gaussian, Exponential, or Laplace. It simultaneously estimates the distribution parameters by optimizing a proper scoring rule, e.g., negative log likelihood (NLL) or Continuous Ranked Probability Score (CRPS). The variant of NGBoost that utilizes Multivariate Gaussian to model multidimensional predictions was presented in \cite{omalley2021multivariate}. RoNGBa \cite{ren2019rongba} is an extension of NGBoost, which improves the performance of NGBoost via a better choice of hyperparameters. This framework has also been adapted to the CatBoost \cite{MalininPU21} with support to only univariate Gaussian distributions, but contrary to the NGBoost, the model outputs all distribution parameters from one model. There is also a group of approaches that were developed in parallel to NGBoost consisting of XGBoostLSS \cite{marz2019xgboostlss} and CatBoostLSS \cite{marz2020catboostlss} which make a connection to well-established statistical framework Generalized Additive Models for Shape, Scale, and Location (GAMLSS) \cite{gamlss}. Like NGBoost, these models use one XGBoost or CatBoost model per parameter, but their training consists of two phases: independent model learning for each parameter and iterative parameter correction. One of the most recent approaches is Probabilistic Gradient Boosting Machine (PGBM) \cite{sprangers2021probabilistic} which treats the leaf weights in each tree as random variables. This approach is capable to model different sets of posterior distributions but is limited to only distributions parameterized with location and scale parameters.

Besides the tree-based probabilistic models, several works investigate the problem of probabilistic regression. In \cite{shao2020conditional} the authors model conditional density estimators for multivariate data with conditional sum-product networks that combines tree-based structures with deep models. In \cite{fakoor2020trade} the authors combine the transformer model with flows for density estimation. The flow models were also applied for future prediction problems in \cite{zikeba2020regflow}. In \cite{sendera2021non} and \cite{maronas2021transforming} the authors propose to integrate flows with Gaussian Processes for probabilistic regression.  

\our{}, to our best knowledge, is the first tree-based model for uni-, and multi-variate probabilistic regression, that is capable to model any distribution for regression outputs. 

\section{Experiments} \label{sec:experiments}
This section evaluates our method on four different setups - univariate regression on synthetic data, univariate regression on mixed-type data, univariate regression on numerical data, and multivariate regression. Our goal is a quantitative and qualitative analysis of \our{} in comparison to the baseline models.

In all experiments, we measure target distribution fit using the negative log likelihood metric in the quantitative part. It is a~natural choice as we expect to deal with heavy-tailed and multimodal distributions. Additionally, we calculate the CRPS metric which is defined as the mean squared difference between the forecasted probabilities and the actual outcomes, over all possible thresholds. It is not the best-suited metric for multimodal distributions, although it is often used for probabilistic forecasting and we would like to understand differences between \our{} and baselines. Moreover, we investigate point estimates that are usually necessary from the application point of view. For that purpose, we use the standard Root Mean Squared Error (RMSE) metric and introduce Root Mean Squared Error at K (RMSE@K) metric. The latter is a version of the RMSE metric that is adjusted for multimodal distributions and takes into account multiple predictions from the model. More details and justifications are provided in the Appendix in sec. \ref{rmseatk}. In the qualitative part, we analyze and discuss the characteristics of obtained probability distributions. Finally, we perform the ablation study whose goal was to justify the design choices. The results of this part are presented in the Appendix in sec. \ref{sec:ablation}.

\subsection{Univariate regression on synthetic data}

This experiment is one of the motivating examples. Here, we want to evaluate the capabilities of \our{} to model data when the true probability distribution is known.

\paragraph{Dataset and methodology} We have created a dataset with two conditioning binary variables. For each possible combination of features, we have proposed different continuous distributions: Normal, Exponential, Mixture of Gaussians, and Gamma (see fig.~\ref{fig:synthetic_data} and details in Appendix, in sec. \ref{sec:app_datasets}). After that, we trained \our{} and CatBoost models. Finally, we calculated negative log likelihood and visualized the obtained probability distributions.

\paragraph{Results} 

After five repetitions of the experiment, we obtained negative log likelihood for CatBoost equal $2.52 \pm 0.01$, and for \our{} equal $2.02 \pm 0.00$. We can observe, that our method effortlessly obtained better results and, contrary to CatBoost, it was able to correctly model all probability distributions (see fig.~\ref{fig:synthetic_data}). This is due to its flexibility in modeling probability distributions resulting from the usage of the CNF component.

\begin{figure}

\begin{center}
    \centering
    \subfigure[\scriptsize{$P(Y|X_1=0, X_2=0)$}]{\includegraphics[width=0.49\columnwidth]{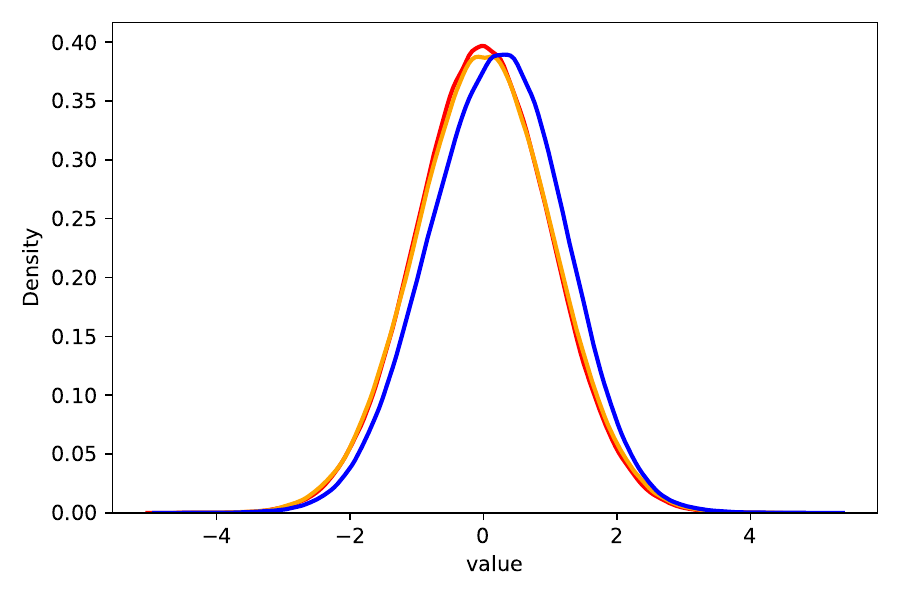}}
    \subfigure[\scriptsize{$P(Y|X_1=0, X_2=1)$}]{\includegraphics[width=0.49\columnwidth]{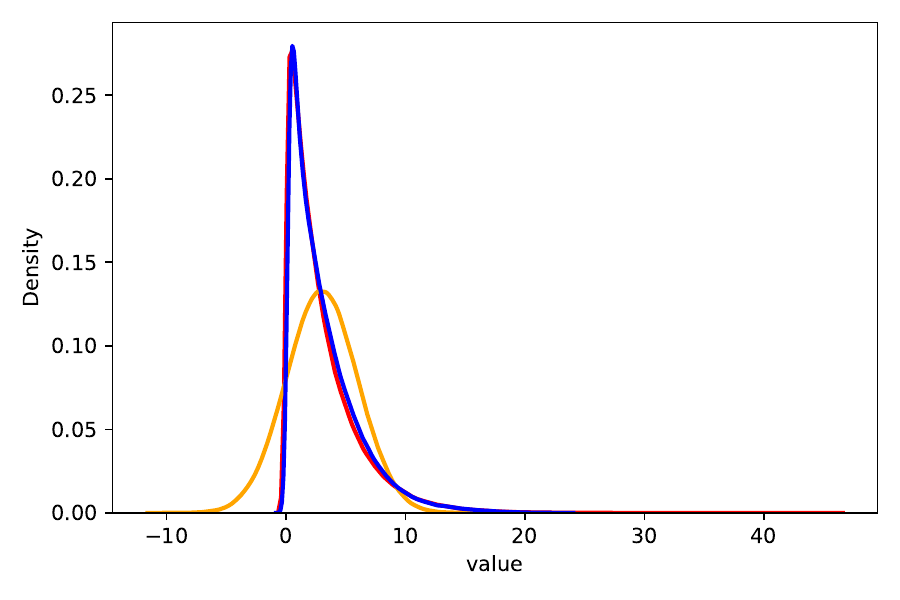}} \\
    \subfigure[\scriptsize{$P(Y|X_1=1, X_2=0)$}]{\includegraphics[width=0.49\columnwidth]{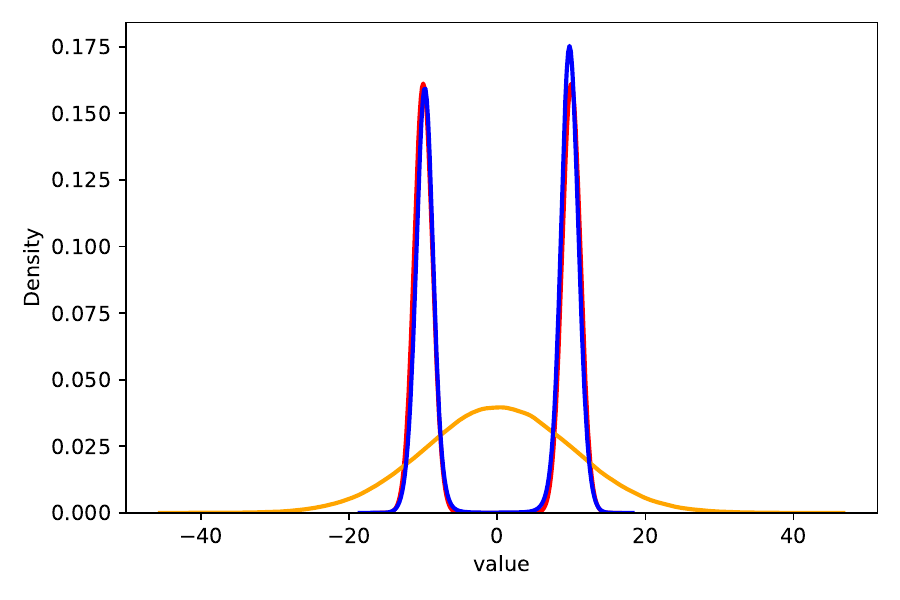}}
    \subfigure[\scriptsize{$P(Y|X_1=1, X_2=1)$}]{\includegraphics[width=0.49\columnwidth]{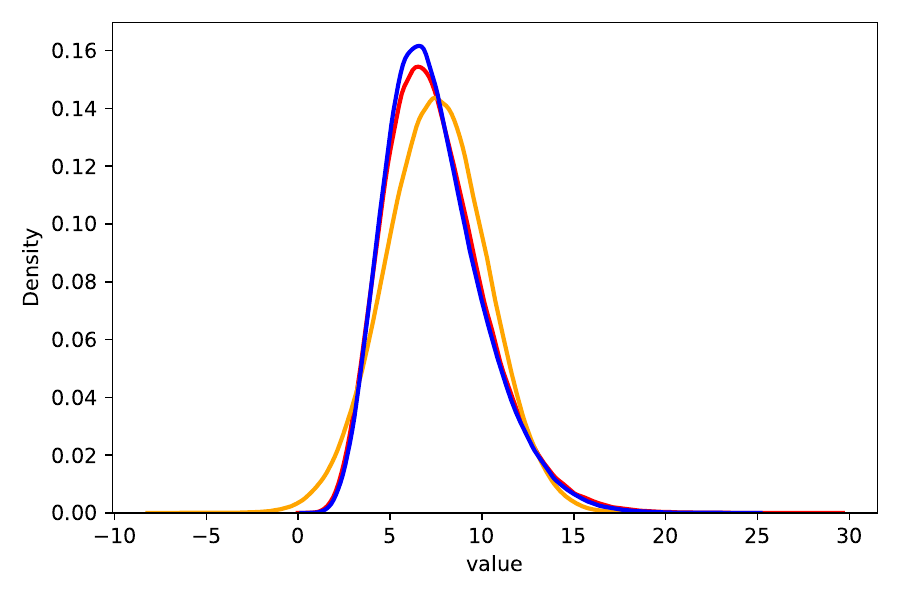}}
    \caption{Comparison of the estimated probability distributions for univariate regression on synthetic data experiment. We can observe that contrary to CatBoost, \our{} was able to correctly model all underlying true probability distributions. Legend: Red - True probability distribution; Blue - \our{}; Orange - CatBoost.}
    \label{fig:synthetic_data}
\end{center}

\end{figure}

\subsection{Univariate regression on mixed-type data}
Our goal is to evaluate and verify our approach to univariate regression problems with mixed-type data. This experiment is the main motivating example of this paper, as tree-based methods cannot model non-gaussian target distributions, and normalizing flows cannot deal with categorical variables without any additional data preparation step.

\paragraph{Datasets and methodology}
To the best of our knowledge, there is no established standard benchmark for regression problems with mixed-type datasets. Thus, we propose seven datasets from the well-known data platform - Kaggle. They have various numbers of samples ranging from a few thousand to a hundred thousand, a different number of categorical and numerical variables. All details of the datasets can be found in tab. \ref{table:catdata_description}.

In terms of the methodology, we follow the standard $80\% / 20\%$ training/testing holdout split. We also split the training dataset to train and validation datasets in the same proportion for the best epoch/iteration selection purposes. All experiments are run 5 times and results are averaged.

For obtaining point estimates from TreeFlow we analyze three approaches: (i) Samples averaging (Avg) - the simple average of samples, (ii) RMSE@1 - usage of the most probable sample, (iii) RMSE@2 - usage of the two most probable samples. Finally, we provide ablation studies regarding the design of the Tree-based Feature Extractor and the Shallow Feature Extractor (see Appendix sec. \ref{sec:ablation}).

\paragraph{Baselines}
Currently, the only approach to work with such problems is a CatBoost which deals out-of-the-box with mixed-type datasets and support modeling target variable with Gaussian distribution. Additionally, we evaluate PGBM with one hot encoding for categorical variables as the representative method for standard tree methods without support for categorical variables. Moreover, this method is also capable of utilizing various parametric distributions. We perform the evaluation on both probabilistic (NLL, CRPS) and deterministic (RMSE / RMSE@K) metrics. 

\paragraph{Results}

\begin{table*}[!ht]
\caption{Comparison of \our{} with existing methods in terms of negative log likelihood (NLL) and Continuous Ranked Probability Score (CRPS) on univariate regression problems with mixed-type data. Our method outperformed both CatBoost and PGBM approaches on most of the datasets thanks to its flexibility in modeling non-gaussian distributions. One Hot Encoding was used for categorical variables for PGBM. Extended information about datasets is provided in tab. \ref{table:catdata_description}.}
\label{table:catdata}

\begin{center}
\begin{footnotesize}
\begin{sc}
\begin{tabular}{l|rrr|rrr}
\toprule
\multirow{2}{*}{Dataset} & \multicolumn{3}{c|}{\textbf{NLL}} & \multicolumn{3}{c}{\textbf{CRPS}} \\
        & CatBoost & PGBM & \our{} & CatBoost & PGBM & \our{} \\
\midrule
Avocado          &      -0.40 $\pm$ 0.01 & -0.45 $\pm$ 0.01 & \textbf{-0.47 $\pm$ 0.03} & 0.0992 $\pm$ 0.0018 & 0.0870 $\pm$ 0.0013 & \textbf{0.0854 $\pm$ 0.0024}  \\
BigMart          &      -0.05 $\pm$ 0.02 & \textbf{-0.10 $\pm$ 0.02} & -0.08 $\pm$ 0.02 & 0.1270 $\pm$ 0.0021 & \textbf{0.1259 $\pm$ 0.0023} & 0.1294 $\pm$ 0.0027  \\
Diamonds         &      -1.80 $\pm$ 0.02 & -1.41 $\pm$ 0.76 &  \textbf{-1.94 $\pm$ 0.03}& 0.0222 $\pm$ 0.0002 & 0.0447 $\pm$ 0.0474 & \textbf{0.0210 $\pm$ 0.0005}  \\
Diamonds 2       &      -1.89 $\pm$ 0.02 & -1.24 $\pm$ 0.83 &  \textbf{-2.14 $\pm$ 0.05}& 0.0217 $\pm$ 0.0002 & 0.0461 $\pm$ 0.0504 & \textbf{0.0197 $\pm$ 0.0005}  \\
Laptop           &      -0.89 $\pm$ 0.08 & \textbf{-0.97 $\pm$ 0.09} & -0.74 $\pm$ 0.13 & 0.0572 $\pm$ 0.0049 & \textbf{0.0474 $\pm$ 0.0034} & 0.0563 $\pm$ 0.0043  \\
Pak Wheel        &      -1.40 $\pm$ 0.05 & -0.53 $\pm$ 0.02 & \textbf{-1.60 $\pm$ 0.03} & 0.0362 $\pm$ 0.0006 & 0.0813 $\pm$ 0.0009 & \textbf{0.0327 $\pm$ 0.0007}  \\
Sydney           &      -0.54 $\pm$ 0.04 &  0.20 $\pm$ 1.02 & \textbf{-0.66 $\pm$ 0.01} & 0.0726 $\pm$ 0.0011 & 0.2383 $\pm$ 0.2646 & \textbf{0.0721 $\pm$ 0.0008}  \\
\bottomrule
\end{tabular}
\end{sc}
\end{footnotesize}
\end{center}

\end{table*}

\begin{table*}
\caption{Comparison of \our{} with existing methods in terms of root mean squared error (RMSE) on univariate regression problems with mixed-type data. \our{} in approach @2 significantly outperforms other baseline methods by taking advantage of multimodal distribution modeling property.}

\begin{center}
\begin{footnotesize}
\begin{sc}
\begin{tabular}{l|rrrrr}
\toprule
\multirow{2}{*}{Dataset} & \multicolumn{5}{c}{\textbf{RMSE}} \\
        & CatBoost & PGBM & \our{}(Avg) & \our{}(@1) & \our{}(@2) \\
\midrule
Avocado          & 0.1939 $\pm$ 0.0043 & \textbf{0.1624 $\pm$ 0.0024} & 0.1676 $\pm$ 0.0058 & 0.1769 $\pm$ 0.0087 & 0.1713 $\pm$ 0.0066 \\
BigMart          & 0.2284 $\pm$ 0.0039 & \textbf{0.2274 $\pm$ 0.0040} & 0.2335 $\pm$ 0.0045 & 0.2514 $\pm$ 0.0087 & 0.2480 $\pm$ 0.0083 \\
Diamonds         & 0.0419 $\pm$ 0.0007 & 0.0403 $\pm$ 0.0006 & 0.0407 $\pm$ 0.0009 & 0.0445 $\pm$ 0.0015 & \textbf{0.0343 $\pm$ 0.0017} \\
Diamonds 2       & 0.0421 $\pm$ 0.0006 & 0.0492 $\pm$ 0.0010 & 0.0398 $\pm$	0.0006 & 0.0460 $\pm$ 0.0014 & \textbf{0.0364 $\pm$	0.0004} \\
Laptop           & 0.1028 $\pm$ 0.0092 & \textbf{0.0848 $\pm$ 0.0063} & 0.1014 $\pm$ 0.0082 & 0.1015 $\pm$ 0.0076 & 0.0958 $\pm$ 0.0058 \\
Pak Wheel        & 0.0783 $\pm$ 0.0009 & 0.1630 $\pm$ 0.0018 & 0.0729 $\pm$ 0.0018 & 0.0796 $\pm$ 0.0021 & \textbf{0.0654 $\pm$ 0.0047} \\
Sydney           & 0.1528 $\pm$ 0.0057 & 0.1561 $\pm$ 0.0047 & 0.1518 $\pm$ 0.0051 & 0.1721 $\pm$ 0.0041 & \textbf{0.1361 $\pm$	0.0066} \\

\bottomrule
\end{tabular}
\end{sc}
\end{footnotesize}
\end{center}
\label{table:catdata_rmse}

\end{table*}

\begin{figure}[h]

\begin{center}
    \centerline{\includegraphics[width=\columnwidth]{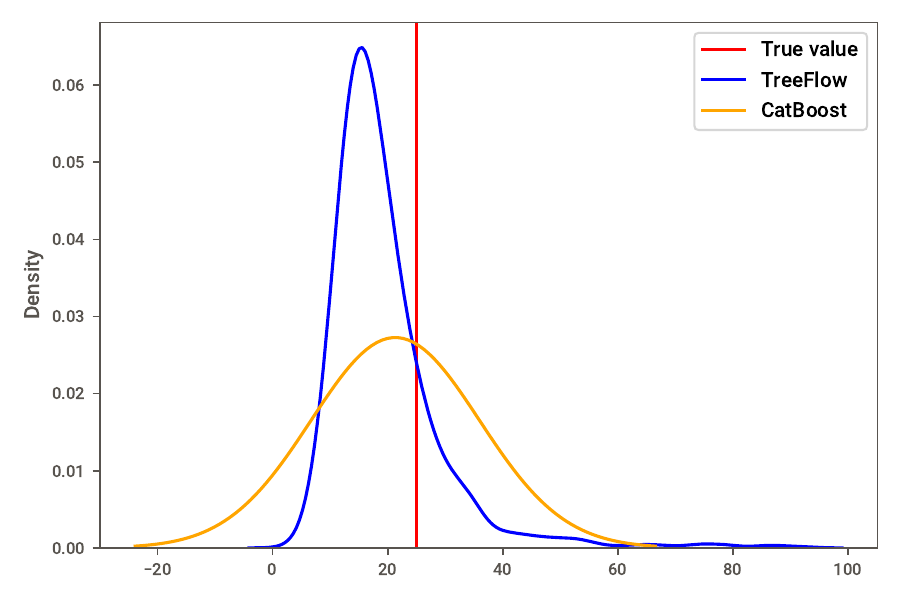}}
    \caption{Estimated probability density functions for the Wine Reviews datasets from the univariate regression on mixed-type data experiment. Both methods predicted similar values for the PDF function; however, the properties of the obtained distributions are entirely different. Contrary to the CatBoost approach, \our{} was able to model heavy-tailed distribution and recognize that negative price values are highly unlikely.}
    \label{fig:wine_reviews}
\end{center}

\end{figure}

The results of the conducted experiments for probabilistic metrics are provided in tab. \ref{table:catdata} and for deterministic metrics in tab. \ref{table:catdata_rmse}. Our method obtains better negative log likelihood scores for most of the datasets and for most of them better CRPS values than reference methods. Furthermore, for the majority of datasets, there is a substantial improvement in the results. In terms of point estimates, \our{} in @2 approach obtains superior results in most of the datasets by the ability to provide multiple predictions for a particular sample that could be modeled by multimodal distributions. The detailed discussion regarding differences between point estimates for \our{} is provided in the Appendix in sec. \ref{rmseatk}
Moreover, we investigated that the target distributions provided by \our{} had more realistic properties such as a heavy tail, multimodality, or does not provide any probability mass for impossible values, e.g., negative values when modeling price as a target variable. The latter example is presented in fig. \ref{fig:wine_reviews}. We analyzed estimated probability density functions for the Wine Reviews datasets for CatBoost and \our{}. Both methods predicted similar values for the PDF function; however, only \our{} was able to model heavy-tailed distribution and recognize that negative price values are highly unlikely.

\subsection{Univariate regression on numerical data} 
We focus on univariate regression problems with only numerical variables in this setup. We aim to evaluate our method on standard probabilistic regression benchmarks in both probabilistic and deterministic approach. Finally, we investigate the properties of the obtained target distributions. 

\paragraph{Datasets and methodology} We use established in the reference methods \cite{DuanADTBNS20, MalininPU21} probabilistic regression benchmark with the exclusion of the Boston dataset due to ethical issues. It contains nine varying-size datasets from the UCI Machine Learning Repository. We follow the same protocol as used in the reference papers. We create 20 random folds for all datasets except Protein (5 folds) and Year MSD (1 fold). We keep $10\%$ of samples as a test set for each of these folds, and the remaining $90\%$ of data we split into an $80\%/20\%$ train/validation for the best epoch selection purposes.

\paragraph{Baselines} We selected four tree-based baseline models: NGBoost \cite{DuanADTBNS20}, RoNGBa \cite{ren2019rongba}, CatBoost \cite{MalininPU21}, PGBM \cite{sprangers2021probabilistic}, and one non tree-based method - Deep Ensemble \cite{Lakshminarayanan17} which was used in \cite{MalininPU21} as a reference method.

\paragraph{Results} 

\begin{table*}
\caption{Comparison of \our{} with existing methods in terms of negative log likelihood (NLL) on univariate regression problems with numerical data. Our method outperformed the other approaches on three datasets and obtained competitive results on others. The superior results were obtained thanks to our method's ability to model multimodal distributions.}

\begin{center}
\begin{scriptsize}
\begin{sc}
\begin{tabular}{l|rrrrrrr}

\toprule
Dataset  & Deep. Ens. & CatBoost & NGBoost & RoNGBa & PGBM & \our{} \\
\midrule
Concrete & 3.06 $\pm$ 0.18 & 3.06 $\pm$ 0.13   & 3.04 $\pm$ 0.17  & 2.94 $\pm$ 0.18 & \textbf{2.75 $\pm$ 0.21} & 3.02 $\pm$ 0.15   \\
Energy   & 1.38 $\pm$ 0.22 & 1.24 $\pm$ 1.28   & 0.60 $\pm$ 0.45  & \textbf{0.37 $\pm$ 0.28} & 1.74 $\pm$ 0.04 & 0.85 $\pm$ 0.35   \\
Kin8nm   & \textbf{-1.20 $\pm$ 0.02} & - 0.63 $\pm$ 0.02 & -0.49 $\pm$ 0.02 & -0.60 $\pm$ 0.03 & -0.54 $\pm$ 0.04  & -1.03 $\pm$ 0.06   \\
Naval    & \textbf{-5.63 $\pm$ 0.05} & -5.39 $\pm$ 0.04  & -5.34 $\pm$ 0.04 & -5.49 $\pm$ 0.04 & -3.44 $\pm$ 0.04  & -5.54 $\pm$ 0.16   \\
Power    & 2.79 $\pm$ 0.04 & 2.72 $\pm$ 0.12   & 2.79 $\pm$ 0.11  & 2.65 $\pm$ 0.08 & \textbf{2.60 $\pm$ 0.02} & 2.65 $\pm$ 0.06   \\
Protein  & 2.83 $\pm$ 0.02 & 2.73 $\pm$ 0.07   & 2.81 $\pm$ 0.03  & 2.76 $\pm$ 0.03 & 2.79 $\pm$ 0.01 & \textbf{2.02 $\pm$ 0.02}  \\
Wine     & 0.94 $\pm$ 0.12 & 0.93 $\pm$ 0.08   & 0.91 $\pm$ 0.06  & 0.91 $\pm$ 0.08 & 0.97 $\pm$ 0.20 & \textbf{-0.56 $\pm$ 0.62} \\
Yacht    & 1.18 $\pm$ 0.21 & 0.41 $\pm$ 0.39   & 0.20 $\pm$ 0.26  & 1.03 $\pm$ 0.44 & \textbf{0.05 $\pm$ 0.28} & 0.72 $\pm$ 0.40  \\
Year MSD & 3.35 $\pm$ NA   & 3.43 $\pm$ NA     & 3.43 $\pm$ NA    & 3.46 $\pm$ NA   & 3.61 $\pm$ NA & \textbf{3.27 $\pm$ NA}   \\

\bottomrule
\end{tabular}
\end{sc}
\end{scriptsize}
\end{center}
\label{table:uci}

\end{table*}

\begin{table*}
\caption{Comparison of \our{} with existing methods in terms of Root Mean Squared Error (RMSE) on univariate regression problems with numerical data.}

\begin{center}
\begin{scriptsize}
\begin{sc}
\begin{tabular}{l|rrrrrrrrr}
\toprule
Dataset  & Deep. Ens.      & CatBoost        & NGBoost         & RoNGBa          & PGBM            & TreeFlow (Avg)  & TreeFlow (@1)   & TreeFlow (@2)   \\
\midrule
Concrete & 6.03 $\pm$ 0.58 & 5.21 $\pm$ 0.53 & 5.06 $\pm$ 0.61 & 4.71 $\pm$ 0.61 & \textbf{3.97 $\pm$ 0.76} & 5.33 $\pm$ 0.65 & 5.41 $\pm$ 0.72 & 5.41 $\pm$ 0.71 \\
Energy   & 2.09 $\pm$ 0.29 & 0.57 $\pm$ 0.06 & 0.46 $\pm$ 0.06 & \textbf{0.35 $\pm $0.07} & \textbf{0.35 $\pm$ 0.06} & 0.64 $\pm$ 0.11 & 0.66 $\pm$ 0.13 & 0.65 $\pm$ 0.12 \\
Kin8nm   & \textbf{0.09 $\pm$ 0.00} & 0.14 $\pm$ 0.00 & 0.16 $\pm$ 0.00 & 0.14 $\pm$ 0.00 & 0.13 $\pm$ 0.01 & \textbf{0.09 $\pm$ 0.00}  & 0.10 $\pm$ 0.01 & 0.10 $\pm$ 0.01 \\
Naval    & \textbf{0.00 $\pm$ 0.00} & \textbf{0.00 $\pm$ 0.00} & \textbf{0.00 $\pm$ 0.00} & \textbf{0.00 $\pm$ 0.00} & \textbf{0.00 $\pm$ 0.00} & \textbf{0.00 $\pm$ 0.00} & \textbf{0.00 $\pm$ 0.00} & \textbf{0.00 $\pm$ 0.00} \\
Power    & 4.11 $\pm$ 0.17 & 3.55 $\pm$ 0.27 & 3.70 $\pm$ 0.22 & 3.47 $\pm$ 0.19 & \textbf{3.35 $\pm$ 0.15} & 3.71 $\pm$ 0.26 & 3.79 $\pm$ 0.26 & 3.79 $\pm$ 0.25 \\
Protein  & 4.71 $\pm$ 0.06 & 3.92 $\pm$ 0.08 & 4.33 $\pm$ 0.03 & 4.21 $\pm$ 0.06 & 3.98 $\pm$ 0.06 & 4.00 $\pm$ 0.27 & 4.79 $\pm$ 0.52 & \textbf{3.01 $\pm$ 0.06} \\
Wine     & 0.64 $\pm$ 0.04 & 0.63 $\pm$ 0.04 & 0.62 $\pm$ 0.04 & 0.62 $\pm$ 0.05 & 0.60 $\pm$ 0.05 & 0.66 $\pm$ 0.05 & 0.73 $\pm$ 0.06 & \textbf{0.41 $\pm$ 0.09} \\
Yacht    & 1.58 $\pm$ 0.48 & 0.82 $\pm$ 0.40 & \textbf{0.50 $\pm$ 0.20} & 0.90 $\pm$ 0.35 & 0.63 $\pm$ 0.21 & 0.75 $\pm$ 0.26 & 0.75 $\pm$ 0.25 & 0.75 $\pm$ 0.26 \\
Year MSD & 8.89 $\pm$ NA   & 8.99 $\pm$ NA   & 8.94 $\pm$ NA   & 9.14 $\pm$ NA   & 9.09 $\pm$ NA   & 9.29 $\pm$ nan  & 10.97 $\pm$ nan & \textbf{8.64 $\pm$ NA}  \\
\bottomrule
\end{tabular}
\end{sc}
\end{scriptsize}
\end{center}
\label{table:uci_rmse}

\end{table*}

The quantitative results for negative log likelihood (NLL) are presented in tab. \ref{table:uci} and for RMSE in tab. \ref{table:uci_rmse}. In terms of the probabilistic metric, our approach outperforms baseline methods on three datasets: Protein, Wine, Year MSD, and obtains competitive results on others. For deterministic metrics, we obtain SOTA results for the same three datasets, and for two (kin8nm and naval) we achieve the same results as the current best methods. 

\begin{figure*}

    \centering
    \subfigure[Multimodal distribution]{\includegraphics[width=0.32\textwidth]{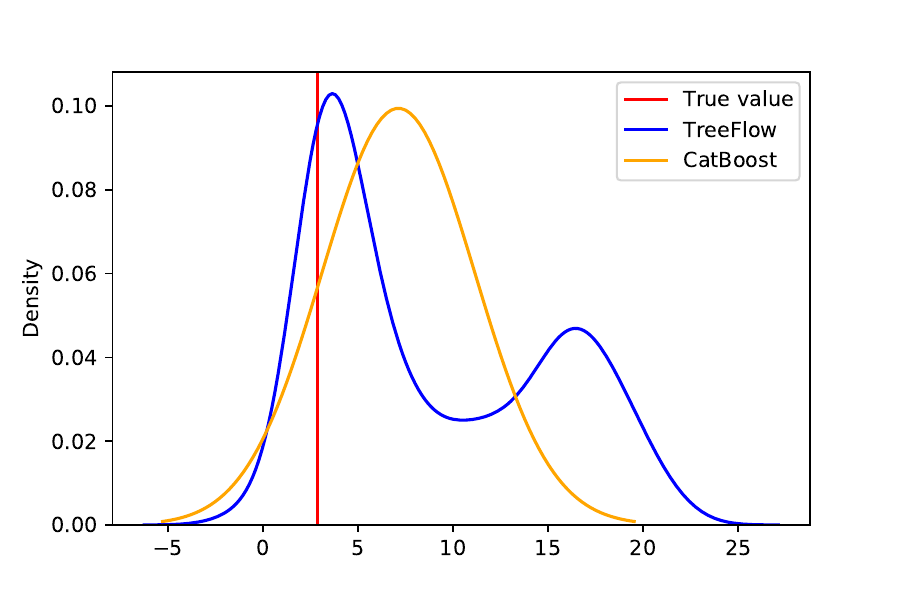}}
    \subfigure[Multimodal, heavy-tailed dist.]{\includegraphics[width=0.32\textwidth]{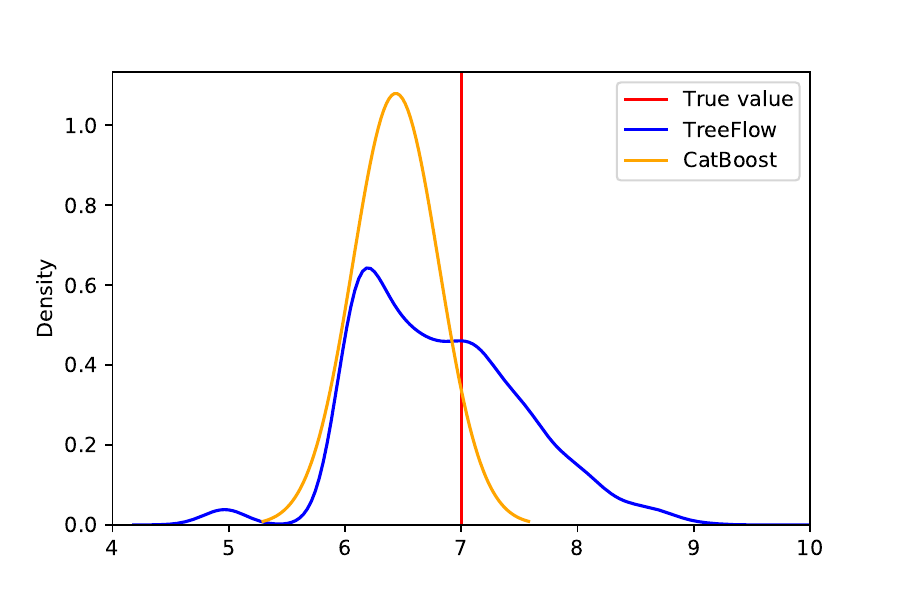}}
    \subfigure[Gaussian distribution]{\includegraphics[width=0.3\textwidth]{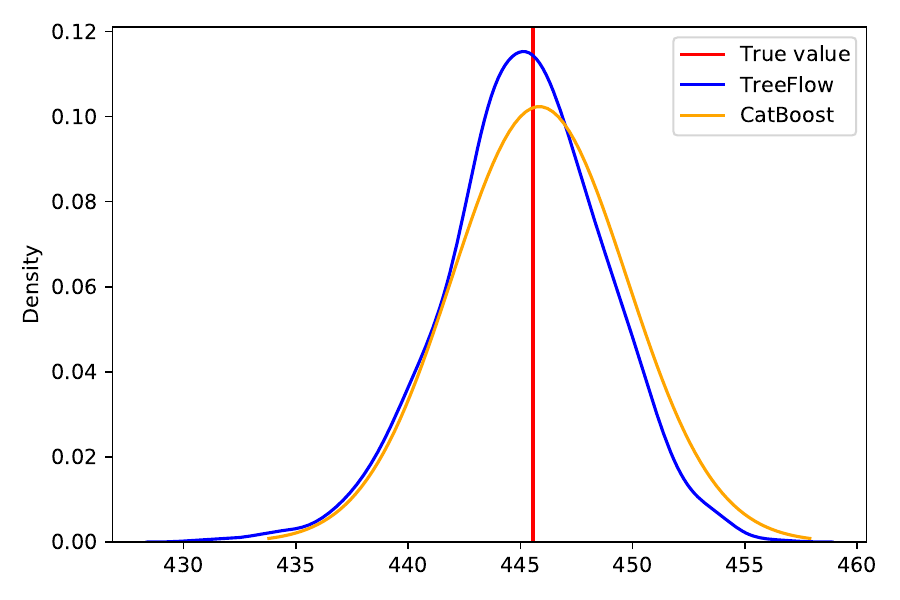}}
    \caption{Estimated probability density functions for three datasets (Protein, Wine Quality, Power Plant) from the univariate regression on numerical data experiment. Depending on the dataset, \our{} is able to model distributions with various properties.}
    \label{fig:uci_plot_summary}
\end{figure*}

To understand the results, we have investigated target distributions. We have compared them with the CatBoost model with default hyperparameters and presented them in fig. \ref{fig:uci_plot_summary}. 

The first subfigure presents results for the Protein dataset. \our{} method has discovered that the underlying target distribution has a bimodal character and was able to correctly estimate the high value of the probability density function for the true value. In contrast, the Gaussian-based method did not have such an ability and incorrectly estimated the center of probability mass between two modes. 
The second subfigure is a representative of naturally occurring integer value datasets: Wine Quality and Year MSD. In this example, our method proposes a multimodal distribution consisting of Gaussian-like and heavy-tailed distributions. Such estimation gives us very rich information for the decision-making process compared to the Gaussian-based approach, which only estimated values around the highest mode and completely ignored information about a minor mode around 5 and a heavy tail for values 8 and 9. 
The last subfigure is a representative example of the rest of the datasets for which our method obtained similar results to baselines. Both methods proposed Gaussian distribution as a target distribution and assuming that this is a correct target distribution, there is no possibility of obtaining significantly better results.

The above-mentioned analysis also explains the results of the deterministic metrics. We incorporated into the decision-making process additional information about the second modality and it resulted in significant gains in prediction accuracy. To the best of our knowledge, it is the first time when these properties were noticed and exploited.

\subsection{Multivariate regression} 
In the last setup, we focus on multivariate regression problems. Our goal is to quantitatively evaluate our method on datasets with various target dimensionality and examine the properties of obtained distributions.

\paragraph{Datasets and methodology} 
Currently, the only tree-based probabilistic multivariate regression problem was approached by \cite{omalley2021multivariate} which proposes a task of two-dimensional oceanographic velocities prediction \cite{michael_o_malley_2021_5644972}. Moreover, we evaluate our method on five more datasets with a broad range of target and feature dimensionality introduced in \cite{YuZTSK21}.

For both groups of datasets, we follow the proposed for these datasets experiment methodology. For the Oceanographic dataset, it is the same protocol as in the univariate regression on numerical data experiment. For the second group, it is a standard training/testing holdout split similar to the univariate regression on the mixed-type data experiment. The exact number of samples is provided in the tab. \ref{table:momogp}.

\paragraph{Baselines}
For this setup, we selected two baseline models. The first approach uses NGBoost, which assumes Multivariate Gaussian distribution and models correlation between target variables. The second approach also uses NGBoost, but the separate model models each target dimension; thus, it assumes independence between targets. We do not consider other Independent Gaussian approaches as they similarly model target distribution.

\paragraph{Results}

\begin{table}
\caption{Comparison of \our{} with existing methods in terms of negative log likelihood on multivariate regression problems. Our method obtains SOTA results on four out of the six datasets thanks to its flexibility in modeling complex distributions. Baseline results for the Oceanographic are taken from the reference paper.}
\begin{center}
\begin{scriptsize}
\begin{sc}
\begin{tabular}{l|rrr}
\toprule
Dataset & Ind NGBoost & NGBoost & \our{} \\ 
\midrule
Parkinsons       &     6.86 &   5.85 &   \textbf{5.26} \\
scm20d           &    94.40 &  94.81 &  \textbf{93.41} \\
WindTurbine      &    -0.65 &  -0.67 &  \textbf{-2.57} \\
Energy           &   \textbf{166.90} & 175.80 & 180.00 \\
usFlight         &     9.56 &   8.57 &   \textbf{7.49} \\ 
\midrule
Oceanographic    &  7.74$\pm$0.02 & \textbf{7.73$\pm$0.02} & 7.84$\pm$0.01 \\

\bottomrule
\end{tabular}
\end{sc}
\end{scriptsize}
\end{center}
\label{table:momogp}
\end{table}

The results of the experiments are provided in tab. \ref{table:momogp}. Our method outperforms baselines by a large margin on three datasets. In contrast to NGBoost-based methods, \our{} was able to capture non-gaussianity in the target distributions. It can be evident on Parkinsons and US Flight datasets where differences between Independent NGBoost and Multivariate NGBoost were significant. They were probably caused by the ability to model the correlation between target variables and \our{} utilized its flexibility to obtain even better results. The other situation is for the Oceanographic dataset, where all results are close. Here, probably true target distribution is similar to the Independent Gaussian distribution; thus, NGBoost and \our{} can not achieve better results. In the last dataset - Energy, the best performing model was Independent NGBoost. We suspect that the high dimensionality of the target distribution was too hard to learn for both NGBoost and TreeFlow methods.

\begin{figure}
\begin{center}
\centerline{\includegraphics[width=\columnwidth]{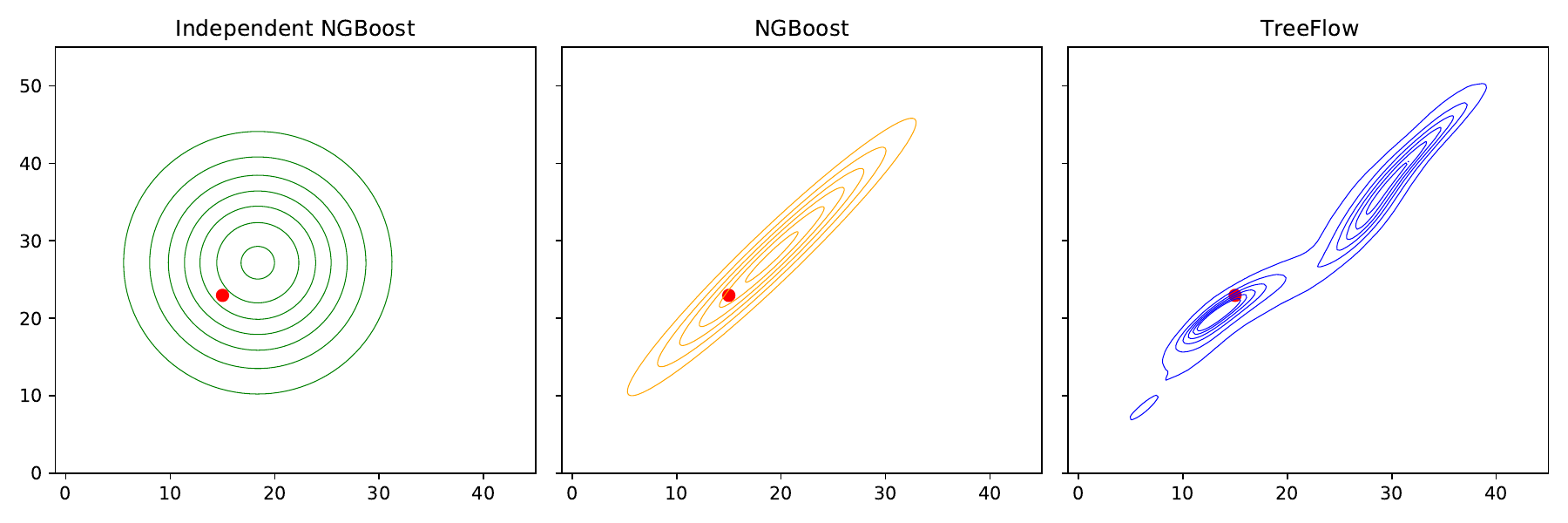}}
\caption{Estimated probability density functions for the Parkinsons datasets from the multivariate regression experiment. \our{} is the most flexible method which enables target distribution to correlate between dimensions and has multiple modes.}
\label{fig:parkinsons}
\end{center}
\end{figure}

Moreover, we investigated target distributions for the Parkinsons dataset. The results of all methods are presented in fig. \ref{fig:parkinsons}. We can easily observe how consecutive methods allow for more flexible distributions. Multivariate NGBoost enables correlation between target variables, while \our{} adds multimodality property.

\section{Conclusions} \label{sec:conclusions}

In this work, we proposed a novel tree-based approach for probabilistic regression. Our method combines the benefits of ensemble decision trees with the capabilities of flow-based models in modeling complex non-Gaussian, multimodal distributions. We evaluated our approach using four experimental settings for both probabilistic and deterministic metrics and achieve SOTA or comparable results on most of them. We also illustrate some properties of \our{} that show the benefits of our approach compared to reference baselines.

\paragraph{Limitations}
The main trade-off introduced by our method is between computational time and the flexibility of the target distribution. Resource-demanding CNF component limits the scalability of our method, but despite this, we were able to deal with datasets of up to half a million observations. Additionally, our method has multiple hyperparameters, which may be challenging to tune in some cases, but we hope that a broad range of experiments provides good intuitions for end users (see sec. \ref{sec:implementation_details}). Lastly, \our{} performs two-staged learning, which might sometimes lead to sub-optimal results. Even so, our method outperforms current baselines, and we hope that \our{} will serve as a strong starting point for future end-to-end approaches.

\paragraph{Broader Impact}
The tree-based models are widely applied in research and industry and often achieve SOTA results. \our{} can be seen as an extension of such models, and all ethical considerations, both positive and negative, regarding regression problems apply to our work. However, as we consider target distribution more complex than parametric, our method can better assess uncertainty in the decision-making process or provide realistic probability distributions (see examples in fig. \ref{fig:wine_reviews}, \ref{fig:uci_plot_summary}, \ref{fig:parkinsons}). Such properties might be crucial, for example, in medicine or finance applications, and have a largely positive societal impact.

\section*{Acknowledgements}
The work conducted by Patryk Wielopolski and Maciej Zieba was supported by the National Centre of Science (Poland) Grant No. 2021/43/B/ST6/02853.

\bibliography{bibliography}

%%%%%%%%%%%%%%%%%%%%%%%%%%%%%%%%%%%%%%%%%%%%%%%%%%%%%%%%%%%%%%%%%%%%%%%%%%%%%%%
%%%%%%%%%%%%%%%%%%%%%%%%%%%%%%%%%%%%%%%%%%%%%%%%%%%%%%%%%%%%%%%%%%%%%%%%%%%%%%%
% APPENDIX
%%%%%%%%%%%%%%%%%%%%%%%%%%%%%%%%%%%%%%%%%%%%%%%%%%%%%%%%%%%%%%%%%%%%%%%%%%%%%%%
%%%%%%%%%%%%%%%%%%%%%%%%%%%%%%%%%%%%%%%%%%%%%%%%%%%%%%%%%%%%%%%%%%%%%%%%%%%%%%%
\newpage
\appendix
\onecolumn

\section{Additional results and discussions}

\subsection{RMSE@K discussion on univariate regression problems with mixed-type data} \label{rmseatk}

\paragraph{RMSE@K Introduction}
For point prediction, we introduce a new metric called root mean squared error at K (RMSE@K). It is a version of the Root Mean Squares Error (RMSE) metric that takes into account multiple predictions from the model. The formula for the metrics is as follows:

$$ \text{RMSE@K(}\mathbf{y}, \mathbf{\hat{y}}\text{)} = \sqrt{\frac{1}{N}\sum_{i=1}^N{\min_{k=1,\ldots, K}(y_i - \hat{y}_{i,k})^2} } $$

It is suited for uni- and multi-variate regression problems where multiple-point predictions are possible. Such a situation origins from probabilistic regression where a model can produce multimodal distributions. In such situations, it's not possible to provide one point estimate, and often for practical reasons, analysis of the whole probabilistic distribution is not feasible. The possible solution is to provide multiple point estimates (usually up to 3 as more modalities rarely occur in real-world settings) for a particular observation.

\paragraph{Detailed results analysis of RMSE@K on univariate regression problems with mixed-type data}
In the experiments, we analyze three approaches for obtaining point estimates from TreeFlow: Samples averaging (Avg) - the simple average of samples, RMSE@1 - usage of the most probable sample, standard RMSE, RMSE@2 - usage of the two most probable samples

In tab. \ref{table:catdata_rmse} we can observe that usually the best results are obtained in the two-point prediction approach, then the samples averaging, and at the end selection of the most probable sample. Such results could be easily explained by considering bimodal distribution that has two almost the same probable modalities. In the first scenario, we are almost always wrong as we do not predict the exact modalities but something between them. In the second scenario, we predict only one modality thus in approximately half of the examples we are right, and in the latter half wrong. In the last scenario, we always select both modalities and check which one is the correct one. Such an approach simulates the real-world scenario where in case of two predictions some end user would check the results and select the correct one.

\subsection{Statistical significance}
Additionally, we have performed the Wilcoxon Signed Rank test with the null hypothesis, that there is no difference between the models' performance, i.e., between \our{} and CatBoost, and between \our{} and PGBM. We used NLL results from univariate regression problems on mixed-type and numeric data (overall 16 datasets). In the first scenario, we obtained a p-value equal to $0.01$, and in the second scenario $0.02$. In both cases it is less than our significance level $\alpha = 0.05$, thus we reject the null hypothesis.

\section{Datasets} \label{sec:app_datasets}

\subsection{Univariate regression on synthetic data}
In order to properly evaluate our method, we need to have a dataset where the true probability distribution is known. It is almost impossible to obtain such a dataset from a real-world scenario; thus, we generated synthetic data. The samples from the dataset were generated using the following probabilistic model:

\begin{small}
\begin{equation}
    \begin{aligned}
    P(Y | X_1 = 0, X_2 = 0) & = \mathcal{N}(Y|\mu = 0, \sigma = 1) \\
    P(Y | X_1 = 0, X_2 = 1) & = \mathcal{E}(Y|\lambda = \tfrac{1}{3}) \\
    P(Y | X_1 = 1, X_2 = 0) & = \tfrac{1}{2}\mathcal{N}(Y|\mu = -10, \sigma = 1) \\
     & + \tfrac{1}{2}\mathcal{N}(Y|\mu = 10, \sigma = 1) \\
    P(Y | X_1 = 1, X_2 = 1) & = \Gamma(Y|k = 7.5, \theta = 1.0)
    \end{aligned}
\end{equation}
\end{small}

The justification for the following probability distribution is as follows. We selected normal distribution to validate if the methods can fit the simplest scenario, exponential distribution to check the fit to heavy-tailed distributions, a mixture of Gaussians for multimodality, and Gamma to check behavior for distributions close to Gaussian distribution.

During the experimental phase for training purposes, we sampled 5,000 observations from each distribution, resulting in a 20,000 samples dataset. We also used 1,000 observations from each distribution for the early stopping / best epoch selection process. The final log likelihood was calculated on a dataset constructed from 500,000 samples per distribution.

\subsection{Univariate regression on mixed-type data}
Extended information about datasets used in univariate regression on mixed-type data experiments is provided in tab.~\ref{table:catdata_description}. We can observe that these datasets consist of various proportions of categorical to numerical features, and cover a broad range of categorical features cardinality. These properties are easily handled by a Tree-based Feature Extractor component with an underlying CatBoost model. Moreover, for selected datasets, we have used log10 transformation of the target variable, mostly due to high absolute values. The non-linearity of this transformation affects the shape of the distribution but favors the CatBoost and PGBM model. Price distribution is usually heavy-tailed, and log transformation makes it more Gaussian. Even though \our{} performed better on these datasets.

\begin{table*}
\caption{Extended information about datasets used in univariate regression problems with the mixed-type experiment. Symbols: $\text{D}_{\text{CAT}}$ - Number of categorical variables, $\text{D}_{\text{NUM}}$ - number of numerical variables, Max card. - Maximum cardinality number among categorical variables, Log transform - Flag if the logarithm of base 10 was used on the target variable.}

\begin{center}
\begin{tiny}
\begin{sc}
\begin{tabular}{l|rrrrrc|c}
\toprule
Dataset & N & D & $\text{D}_{\text{CAT}}$ & $\text{D}_{\text{NUM}}$ & Max card. & Log transform & Link\\ 
\midrule
Avocado          &   18,249 & 11 & 3 & 8 & 54  & \xmark  & Link\tablefootnote{\url{https://www.kaggle.com/neuromusic/avocado-prices}} \\
BigMart          &    8,523 & 10 & 6 & 4 & 16  & \cmark & Link\tablefootnote{\url{https://www.kaggle.com/yasserh/bigmartsalesdataset}} \\
Diamonds         &   53,940 &  9 & 3 & 6 & 8   & \cmark & Link\tablefootnote{\url{https://www.kaggle.com/shivam2503/diamonds}} \\
Diamonds 2       &  119,307 &  7 & 6 & 1 & 10  & \cmark & Link\tablefootnote{\url{https://www.kaggle.com/miguelcorraljr/brilliant-diamonds}} \\
Laptop           &    1,303 & 10 & 7 & 3 & 118 & \cmark & Link\tablefootnote{\url{https://www.kaggle.com/muhammetvarl/laptop-price}} \\
Pak Wheel        &   76,690 &  7 & 4 & 3 & 326 & \cmark & Link\tablefootnote{\url{https://www.kaggle.com/ebrahimhaquebhatti/75000-used-cars-dataset-with-specifications}} \\
Sydney Housing   &  199,504 &  6 & 3 & 3 & 685 & \cmark & Link\tablefootnote{\url{https://www.kaggle.com/mihirhalai/sydney-house-prices}} \\
\bottomrule
\end{tabular}
\end{sc}
\end{tiny}
\end{center}
\label{table:catdata_description}

\end{table*}

\subsection{Univariate regression on numerical data}
Datasets for univariate regression on numerical data experiments were used without any preprocessing. None of the datasets contained missing values. Extended information regarding dataset sizes is provided in tab. \ref{table:uci_description}.

\begin{table}
\caption{Extended information about datasets used in univariate regression problems with numerical data.}

\begin{center}
\begin{small}
\begin{sc}
\begin{tabular}{l|rrr}
\toprule
Dataset  & N      & D   & CV Splits \\ 
\midrule
Concrete & 1030   & 8  & 20 \\
Energy   & 768    & 8  & 20 \\
Kin8nm   & 8192   & 8  & 20 \\
Naval    & 11934  & 16 & 20 \\
Power    & 9568   & 4  & 20 \\
Protein  & 45730  & 9  & 5  \\
Wine     & 1588   & 11 & 20 \\
Yacht    & 308    & 6  & 20 \\
Year MSD & 515345 & 90 & 1  \\ 
\bottomrule
\end{tabular}
\end{sc}
\end{small}
\end{center}
\label{table:uci_description}

\end{table}

\subsection{Multivariate regression}
Datasets for multivariate regression experiment were used without any preprocessing except Oceanographic. Here, we multiplied the target value by 100 due to numerical stability. The same operation was used in the reference paper. Moreover, none of the datasets contained missing values. Additional information about dataset sample sizes, dimensionalities of features, and target variables are presented in tab. \ref{table:momogp_description}.

\begin{table}
\caption{Extended information about datasets used in multivariate regression problems.}

\begin{center}
\begin{scriptsize}
\begin{sc}
\begin{tabular}{l|rrrr}
\toprule

Dataset & $\text{N}_{\text{train}}$ & $\text{N}_{\text{test}}$ & D & P  \\ 

\midrule
Parkinsons       & 4,112   & 1,763   & 16 & 2   \\
scm20d           & 7,173   & 1,793   & 61 & 16  \\
WindTurbine      & 4,000   & 1,000   &  8 & 6   \\
Energy           & 57,598  & 14,400  & 32 & 17  \\
usFlight         & 500,000 & 200,000 &  8 & 2   \\ 
\midrule
Oceanographic    & 414,697 &  20 CV  &  9 & 2   \\

\bottomrule
\end{tabular}
\end{sc}
\end{scriptsize}
\end{center}
\label{table:momogp_description}

\end{table}

\section{Implementation details} \label{sec:implementation_details}
In this section, we cover the essential information related to implementation details. The code for experiments is available in the Supplementary Materials. After the review process, they will be published publicly on the GitHub repository.

We presented the architecture of \our{} model in fig. \ref{fig:architecture}. We have used CatBoost as the Tree-based Feature Extractor, one layer neural network with tanh activation function as Shallow Feature Extractor and Conditional Continuous Normalizing Flow presented in \cite{yang2019pointflow} for the Conditional CNF component. 

In terms of the multipoint estimation, it starts with sampling 1000 observations from the target distribution. In the next step, we use kernel density estimation (KDE) that approximates the probability density function (PDF), and then we run the find\_peaks procedure provided by the SciPy Python package. As a side note, \our{} provides PDF, however, the function is highly unsmooth and our practical experiments showed that KDE approximation works much better.

In the tab. \ref{table:hp_search_uci} and tab. \ref{table:hp_search_momogp} we used the following naming convention:

\begin{itemize}
    \item Depth - maximum depth of the single tree in the CatBoost ensemble;
    \item N trees - number of trees in the CatBoost ensemble;
    \item Context dim - dimensionality of the output layer of the Shallow Feature Extractor;
    \item Hidden dim - dimensionality of the consecutive layers in the dynamic function of the CNF block;
    \item N blocks - number of CNF blocks;
    \item N epochs - number of training epochs.
\end{itemize}

For all experiments purposes, we used a machine with AMD Ryzen 9 5950X 16-Core Processor CPU, 2 NVIDIA GeForce 2080 Ti GPUs, and 64 GB RAM. 

\subsection{Univariate regression on synthetic data}
In this experiment, we were responsible for training both CatBoost and \our{} models. CatBoost was trained using default hyperparameters as they are known to work very well out of the box. For \our{} we used: Depth: 2, N trees: 100, Context dim: 50, Hidden dims: [50, 10], N blocks: 2, Num epochs: 50.

\subsection{Univariate regression on mixed-type data}
For the reason of fair comparisons, in this experiment, we used the set of the same hyperparameters for all datasets. We were responsible for training all methods. The default hyperparameters of CatBoost and PGBM were used, and for \our{} we used Depth: 4, N trees: 200, Context dim: 128, Hidden dim: [16, 16], N blocks: 2, Num epochs: 150.

% \begin{table*}
% \caption{Hyperparameters used for \our{} method in the univariate regression on mixed-type data experiment.}
% 
% \begin{center}
% \begin{small}
% \begin{sc}
% \begin{tabular}{l|rr|rrr|r}
% \toprule

% \multirow{2}{*}{Dataset} & \multicolumn{2}{|c|}{Tree parameters} & \multicolumn{3}{|c|}{Flow parameters} & \multicolumn{1}{c}{General} \\
%                          & Depth & N trees                & Context Dim & Hidden Dim & N blocks  & N epochs\\
% \midrule
% Avocado   & 4 & 200 & 100 & [200, 200, 100, 50] & 5 & 50 \\
% BigMart   & 2 & 500 & 100 & [200, 200, 100, 50] & 5 & 50 \\
% Diamonds  & 4 & 200 & 100 & [200, 200, 100, 50] & 5 & 30 \\
% Diamonds 2& 4 & 200 & 100 & [100, 100, 50] & 5 & 30 \\
% Laptop    & 1 & 500 & 100 & [100, 100, 50] & 3 & 100 \\
% Pak Wheels & 5 & 100& 100 & [100, 100, 50] & 3 & 50 \\
% Sydney Housing& 4 & 200 & 100 & [200, 200, 100, 50] & 5 & 30 \\
% Wine Reviews& 3 & 100 & 100 & [100, 100, 50] & 3 & 10 \\

% \bottomrule
% \end{tabular}
% \end{sc}
% \end{small}
% \end{center}
% \label{table:hp_search_catdata}
% 
% \end{table*}

\subsection{Univariate regression on numerical data}
For this experiment, we performed a hyperparameter search. The process consisted of both grid search and manual trials and errors, i.e., we performed an initial grid search to obtain intuitions on the validation dataset and then ran consecutive grid searches with changed ranges of hyperparameters. The final hyperparameter setting is presented in tab. \ref{table:hp_search_uci}. Results in tab. \ref{table:uci} for the reference methods were taken from the reference papers \cite{DuanADTBNS20, MalininPU21} except PGBM that was trained by us with hyperparameters from \cite{sprangers2021probabilistic}.

\begin{table*}
\caption{Hyperparameters used for \our{} method in the univariate regression on numerical data experiment. Outer square brackets are equivalent to a grid search among provided values.}

\begin{center}
\begin{scriptsize}
\begin{sc}
\begin{tabular}{l|rr|rrr|r}
\toprule
\multirow{2}{*}{Dataset} & \multicolumn{2}{|c|}{Tree parameters} & \multicolumn{3}{|c|}{Flow parameters} & \multicolumn{1}{c}{General} \\
                         & Depth & N trees                & Context Dim & Hidden Dim & N blocks  & N epochs\\
\midrule
Concrete & 1 & [300, 500, 750] & [50, 100, 200] & [[100, 100, 50], [200, 100, 100, 50]] & 3 & 100 \\
Energy   & [1, 2] & [100, 300     ] & [40, 100     ] & [[80, 40], [80, 80, 40], [80, 80, 80, 40]] & 3 & 200 \\
Kin8nm   & [1, 2] & [100, 300     ] & [40, 100     ] & [[80, 40], [80, 80, 40], [80, 80, 80, 40]] & 3 & 20 \\
Naval    & 4 & [500, 750] & 200 & [[100, 100, 50], [200, 100, 100, 50]] & 4 & 25 \\
Power    & 4 & 500 & [100, 200] & [[100, 50], [100, 100, 50], [200, 100, 100, 50]] & [3, 5] & 30 \\
Protein  & 4 & 750 & 100 & [100, 100, 50] & 3 & 25 \\
Wine     & [1, 2] & [100, 300] & [40, 100     ] & [[80, 40], [80, 80, 40]] & 3 & 200 \\
Yacht    & [1, 2] & [300, 500, 750] & [50, 100, 200] & [[100, 100, 50], [200, 100, 100, 50]] & [1, 2] & 100 \\
Year MSD & [1, 2] & [100, 300] & [40, 100     ] & [[80, 40], [80, 80, 40], [80, 80, 80, 40]] & 3 & 3 \\

\bottomrule
\end{tabular}
\end{sc}
\end{scriptsize}
\end{center}
\label{table:hp_search_uci}

\end{table*}

\subsection{Multivariate regression}
In this experiment, we trained models with hyperparameter search for both NGBoost and \our{} models. For NGBoost models (Independent and Multivariate Gaussian), we performed a grid search on all datasets except Oceanographic, where results work taken from the reference paper \cite{omalley2021multivariate}. The range of the hyperparameters was inspired from \cite{DuanADTBNS20}. The parameters are presented in tab. \ref{table:hp_search_momogp_ngboost}. For the \our{} method, the same approach was applied with the final hyperparameter space presented in tab. \ref{table:hp_search_momogp}.

\begin{table*}[ht!]
\caption{Hyperparameters used for NGBoost methods in the multivariate regression on
experiment. Square brackets are equivalent to a grid search among provided values.}

\begin{center}
\begin{small}
\begin{sc}
\begin{tabular}{l|rrr|r}
\toprule

\multirow{2}{*}{Dataset} & \multicolumn{3}{|c|}{Tree parameters} & NGBoost parameters \\
                         & Max Depth & Max Leaf Nodes & Min Samples Leaf & Num trees\\
\midrule
Parkinsons    & [5, 10, 15] & [8, 15, 32, 64] & [1, 15, 32] & [100, 300, 500] \\
Scm20d        & 15 & [8, 15, 32, 64] & [1, 15, 32] & [100, 300, 500] \\
WindTurbine   & [5, 10, 15] & [8, 15, 32, 64] & [1, 15, 32] & [100, 300, 500] \\
Energy        & 15 & [8, 15, 32, 64] & [1, 15, 32] & [100, 300, 500] \\
UsFlight      & [5, 10, 15] & [8, 15, 32, 64] & [1, 15, 32] & [100, 300, 500] \\

\bottomrule
\end{tabular}
\end{sc}
\end{small}
\end{center}
\label{table:hp_search_momogp_ngboost}

\end{table*}

\begin{table*}[ht!]
\caption{Hyperparameters used for \our{} method in the multivariate regression experiment. Outer square brackets are equivalent to grid search among provided values.}

\begin{center}
\begin{scriptsize}
\begin{sc}
\setlength{\tabcolsep}{4pt}
\begin{tabular}{l|rr|rrr|r}
\toprule

\multirow{2}{*}{Dataset} & \multicolumn{2}{|c|}{Tree parameters} & \multicolumn{3}{|c|}{Flow parameters} & \multicolumn{1}{c}{General} \\
                         & Depth & N trees                & Context Dim & Hidden Dim & N blocks  & N epochs\\
\midrule
Parkinsons   & [1, 2] & [100, 300] & [40, 100] & [[80, 40], [80, 40, 40], [80, 80, 80, 40]]  & 3 & 500 \\
Scm20d       & [1, 2] & [100, 300] & [40, 100] & [[80, 40], [80, 40, 40], [80, 80, 80, 40]]  & 3 & 200 \\
WindTurbine  & 1 & [500, 750] & [50, 100] & [[100, 50], [100, 100, 50], [200, 100, 100, 50]] & [3, 5] & 150 \\
Energy       & [1, 2] & [100, 300] & [40, 100] & [[80, 40], [80, 40, 40], [80, 80, 80, 40]]  & 3 & 30 \\
UsFlight     & [1, 2] & [100, 300, 500] & [40, 80, 120] & [[80, 40, 40], [80, 80, 80, 40], [200, 100, 100, 50]]  & [1, 3, 5] & 5 \\
Oceanographic& 2 & [750, 1000] & [100] & [[50, 50]] & 1 & 30 \\

\bottomrule
\end{tabular}
\end{sc}
\end{scriptsize}
\end{center}
\label{table:hp_search_momogp}

\end{table*}

\section{Ablation study} 
\label{sec:ablation}
In this section, we perform two experiments to analyze the contribution of specific \our{}'s components to the overall performance. The first experiment discusses the Tree-based Feature Extractor component and the second Shallow Feature Extractor.

\subsection{Tree-based Feature Extractor} \label{sec:ablation_ohe}
We introduced the Tree-based Feature Extractor component as the tree-based model, and in the experiments, we specifically focused on the CatBoost implementation. Our motivation was to enable our method to deal with categorical variables efficiently. The most common alternative is to use One Hot Encoder, which encodes each possible category as a binary vector of category occurrence.

In this ablation study, we performed an experiment where we replaced CatBoost with One Hot Encoder as a Feature Extractor. In practice, it reduces the model to CNF with an additional MLP layer for the conditioning factor encoding (in this work called Shallow Feature Extractor). The experiment methodology and hyperparameters were the same as in the Univariate regression on mixed-type data experiment. The results of the experiments are presented in tab.~\ref{table:ablation_ohe}. We can observe that for almost all of the datasets \our{} obtains better or comparable results to CNF and thus, we conclude that the Tree-based Feature Extractor is a crucial component of \our{}.

\begin{table*}
\caption{Results of the ablation study regarding Tree-based Feature Extractor. In this scenario Tree-based Feature Extractor (CatBoost backbone) in \our{} was replaced by the One Hot Encoder (OHE) which results in CNF model. $\text{D}_{\text{OHE}}$ represents the number of features in the dataset after One Hot Encoding categorical variables.}

\begin{center}
\begin{small}
\begin{sc}
\begin{tabular}{l|ll|rr}
\toprule
Dataset        & D  & $\text{D}_{\text{OHE}}$ & \our{} & CNF (\our{} with OHE) \\
\midrule
Avocado        & 11 & 65                      & \textbf{-0.47 $\pm$ 0.03} & -0.27 $\pm$ 0.02 \\
BigMart        & 10 & 46                      & -0.08 $\pm$ 0.02 & \textbf{-0.12 $\pm$ 0.01} \\
Diamonds       & 9  & 26                      & \textbf{-1.94 $\pm$ 0.03} & -1.78 $\pm$ 0.03 \\
Diamonds 2     & 7  & 37                      & \textbf{-2.14 $\pm$ 0.05} & -1.53 $\pm$ 0.13 \\
Laptop         & 10 & 344                     & \textbf{-0.74 $\pm$ 0.13} & -0.70 $\pm$ 0.27 \\
Pak Wheel      & 7  & 402                     & \textbf{-1.60 $\pm$ 0.03} & -1.26 $\pm$ 0.02 \\
Sydney Housing & 6  & 697                     & \textbf{-0.66 $\pm$ 0.01} & -0.60 $\pm$ 0.06 \\
\bottomrule
\end{tabular}
\end{sc}
\end{small}
\end{center}
\label{table:ablation_ohe}

\end{table*}

\subsection{Shallow Feature Extractor} \label{sec:ablation_sfe}
The next component which we introduced was the Shallow Feature Extractor. Its task was to map high-dimensional binary vectors extracted from forest structures to low-dimensional feature representation. The main goal of that operation was to reduce computational overhead. We present an ablation study where we exclude the Shallow Feature Extractor component and pass the output of the Tree-based Feature Extractor directly to the Conditional CNF component.

We used the same methodology as in the previous ablation study, but we only calculate the training time of one epoch of the model. The experiment was run on a CPU, so the relative speed up is the crucial factor in the comparison. The results are presented in tab. \ref{table:ablation_sfe}. We can observe that \our{} was on average 11 times faster than a model without Shallow Feature Extractor.

Concluding, our ablation study showed that Shallow Feature Extractor is a crucial component of the method in terms of computational time performance.

\begin{table*}
\caption{Results of the ablation study regarding Shallow Feature Extractor. In this scenario Shallow Feature Extractor (SFE) in \our{} was not present. $\text{D}_{\text{SFE}}$ represented the number of features extracted from the Tree-based Feature Extractor and passed directly to the Conditional CNF component. The calculation time was the one epoch training time and the experiments were conducted on CPU time. \our{} on average was 11 times faster than the model without SFE.}

\begin{center}
\begin{small}
\begin{sc}
\begin{tabular}{l|ll|rr|rr|r}
\toprule
Dataset        & D  & $\text{D}_{\text{SFE}}$ & \our{} & \our{} without SFE & Speed up \\
\midrule
Avocado        & 11 & 1600                    &  8.87 $\pm$ 0.19 s & 113.25 $\pm$ 2.01 s & 12.7 x \\
BigMart        & 10 & 1600                    &  4.67 $\pm$ 0.13 s &  41.37 $\pm$ 0.79 s &  8.9 x \\
Diamonds       & 9  & 1600                    & 25.26 $\pm$ 0.33 s & 345.28 $\pm$ 5.03 s & 13.7 x \\
Diamonds 2     & 7  & 1600                    & 55.56 $\pm$ 0.28 s & 756.62 $\pm$ 17.91 s& 13.6 x \\
Laptop         & 10 & 1600                    &  1.90 $\pm$ 0.08 s &   5.84 $\pm$ 0.07 s &  3.1 x \\
Pak Wheel      & 7  & 1600                    & 36.20 $\pm$ 0.89 s & 493.18 $\pm$ 7.82 s & 13.6 x \\
Sydney Housing & 6  & 1600                    & 91.21 $\pm$ 1.69 s &1192.03 $\pm$ 30.09 s& 13.1 x \\
\bottomrule
\end{tabular}
\end{sc}
\end{small}
\end{center}
\label{table:ablation_sfe}

\end{table*}

\end{document}